\pdfoutput=1

\documentclass[11pt]{article}

\usepackage{acl}

\usepackage{times}
\usepackage{latexsym}

\usepackage{graphicx}
\usepackage{subfig}

\usepackage{float}

\usepackage{url}
\usepackage{amsmath}
\usepackage{tabularx} 
\usepackage{booktabs}
\usepackage[T1]{fontenc}

\usepackage[utf8]{inputenc}

\usepackage{microtype}

\usepackage{inconsolata}

\title{Application of LLM Agents in Recruitment: A Novel Framework for Resume Screening}

\author{
  Chengguang Gan\textsuperscript{1} \quad
  Qinghao Zhang\textsuperscript{2} \quad
  Tatsunori Mori\textsuperscript{1} \\
  \textsuperscript{1}Yokohama National University, Japan \\
  \texttt{gan-chengguan-pw@ynu.jp, tmori@ynu.ac.jp} \\
  \textsuperscript{2}Department of Information Convergence Engineering, \\
  Pusan National University, South Korea \\
  \texttt{zhangqinghao@pusan.ac.kr}
}

\begin{document}
\maketitle
\begin{abstract}
The automation of resume screening is a crucial aspect of the recruitment process in organizations. Automated resume screening systems often encompass a range of natural language processing (NLP) tasks. This paper introduces a novel Large Language Models (LLMs) based agent framework for resume screening, aimed at enhancing efficiency and time management in recruitment processes. Our framework is distinct in its ability to efficiently summarize and grade each resume from a large dataset. Moreover, it utilizes LLM agents for decision-making. To evaluate our framework, we constructed a dataset from actual resumes and simulated a resume screening process. Subsequently, the outcomes of the simulation experiment were compared and subjected to detailed analysis. The results demonstrate that our automated resume screening framework is 11 times faster than traditional manual methods. Furthermore, by fine-tuning the LLMs, we observed a significant improvement in the F1 score, reaching 87.73\%, during the resume sentence classification phase. In the resume summarization and grading phase, our fine-tuned model surpassed the baseline performance of the GPT-3.5 model \cite{ouyang2022training}. Analysis of the decision-making efficacy of the LLM agents in the final offer stage further underscores the potential of LLM agents in transforming resume screening processes.
\end{abstract}

\section{Introduction}

\begin{figure}[!t]
\centering
\includegraphics[width=228 pt]{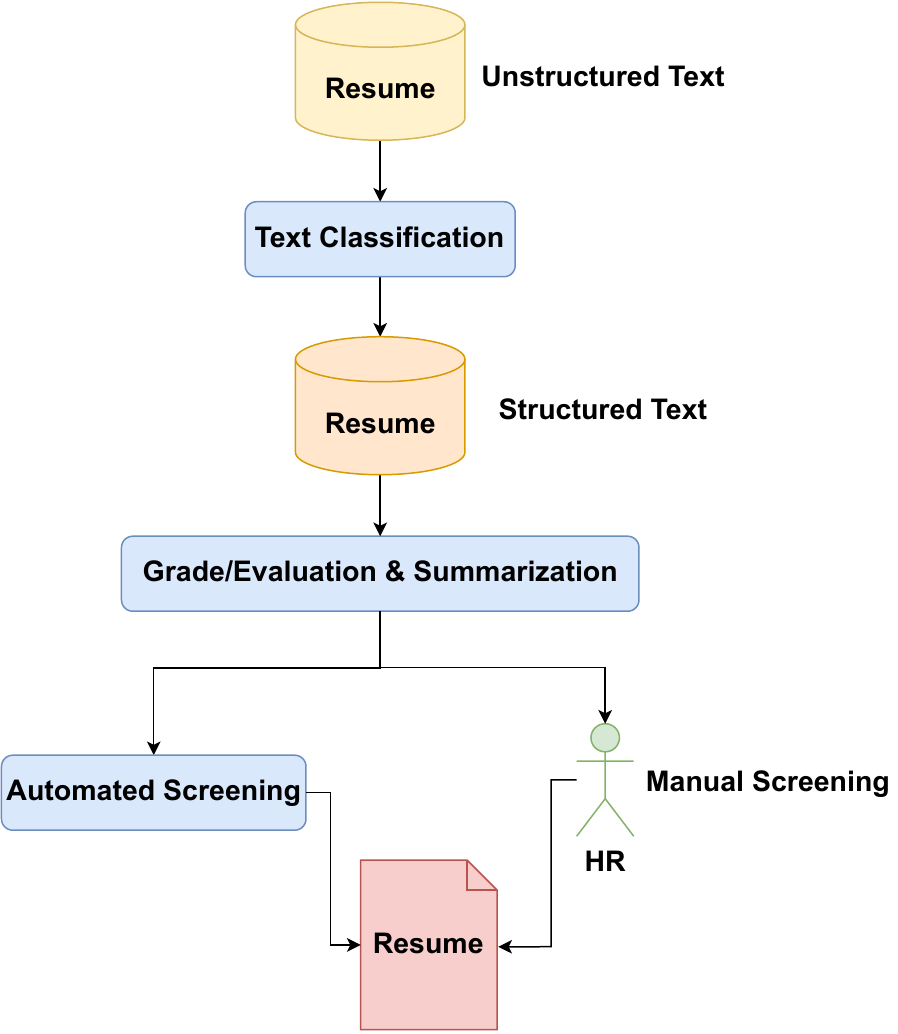}
\caption{\label{1figure1}The Process of automated resume screening.}

\end{figure}

Resume screening is a crucial aspect of recruitment for all companies, particularly larger ones, where it becomes a labor-intensive and time-consuming endeavor. In contrast to smaller firms, a large corporation might receive thousands of resumes during a hiring phase, making efficient screening of these numerous applications a significant challenge. To reduce labor costs associated with resume screening, developing an automated framework is essential. Utilizing natural language processing (NLP) technology for this purpose is increasingly becoming the preferred approach.

The automated resume screening \cite{Singh2010PROSPECTAS} process encompasses two primary components: information extraction \cite{singhal2001modern} and evaluation. As illustrated in Figure \ref{1figure1}, resumes typically exist as unstructured or semi-structured text, varying in format. The initial step of the automated framework is to convert this unstructured text into a structured format. This process involves a key NLP task: text classification \cite{10.1145/3544558}, specifically sentence classification \cite{10.1145/3439726}. It entails extracting and classifying sentences related to personal information, education, and work experience, transforming them into structured data that is easily stored and manipulated.

Upon structuring the resume text, it must then be summarized  and evaluated. The lower part of Figure \ref{1figure1} depicts this process, which includes both automatic and manual screening. Manual screening involves grading and summarizing extensive sections of the resume text, after which the graded and summarized resumes are presented to HR for review, leading to the selection of qualified candidates. This approach significantly reduces the time HR personnel spend perusing resumes and deliberating decisions by shortening the resume text and implementing a grading system for ranking. The aim is to enhance the efficiency of the screening process. NLP technology can also automate this process, culminating in the output of qualified resumes.

\begin{figure}[!h]
\centering
\includegraphics[width=228 pt]{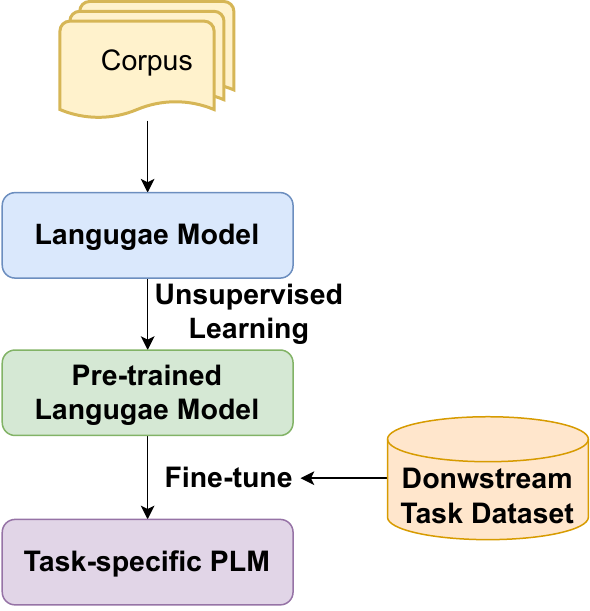}
\caption{\label{2figure2}The illustration reprehsents the process of pre-training a language model and applying the pre-trained language model to a downstream task through fine-tuning method.}

\end{figure}

In the preceding discussion, we elucidated two NLP tasks pertinent to the automated extraction of information from resumes. Addressing these tasks necessitates the employment of Language Models (LMs) . Presently, the most prevalent infrastructure for LMs is the transformer architecture \cite{vaswani2017attention}, distinguished by its attention mechanism. These LMs are predominantly trained on extensive corpora, endowing them with a broad spectrum of knowledge. The seq2seq (sequence-to-sequence) \cite{sutskever2014sequence} structure is instrumental in this context, enabling the conversion of an input sequence into a predicted output sequence. This mechanism facilitates the adaptability of LMs to a diverse range of NLP tasks.

As illustrated in Figure \ref{2figure2}, the process of LMs spans from their training to their application in various downstream NLP tasks. The initial phase involves assembling a substantial corpus for unsupervised learning, encompassing a broad array of general knowledge. This corpus is typically derived from sources such as Wikipedia \footnote{https://www.wikipedia.org/} and extensive web content. Subsequently, these voluminous, unlabeled corpora serve as the foundation for training LMs. Through this process, LMs acquire foundational linguistic competencies and general knowledge autonomously. Following the pre-training phase, Pre-trained Language Models (PLMs) \cite{min2023recent} undergo fine-tuning \cite{ding2023parameter} with different datasets tailored to specific downstream tasks. The culmination of this process is the development of task-specific PLMs, capable of effectively predicting or processing relevant NLP tasks.

The initial PLMs, such as BERT \cite{devlin2018bert}, T5 \cite{raffel2020exploring}, and GPT-2 \cite{radford2019language}, were characterized by their relatively modest size, containing only several hundred million parameters. However, the advent of GPT-3 \cite{brown2020language} marked a significant leap in this field, boasting an impressive 135 billion parameters. This escalation was not merely quantitative but also qualitative, as evidenced by the subsequent development of ChatGPT \cite{ouyang2022training}. ChatGPT underscored how expanding the pre-trained corpus and increasing the parameter count of PLMs could substantially enhance their capabilities, thereby heralding a new era in the development of Large Language Models (LLMs) \cite{zhao2023survey}.

Despite these advancements, concerns have arisen regarding the closed-source models developed by major corporations, particularly in terms of user security. The primary issue lies in the potential for private information leakage. Utilizing these LLMs typically requires users to upload their data, creating a risk of data compromise. This is especially pertinent in applications like resume screening, where sensitive personal information is involved. In contrast to closed-source models like GPT-3.5 and GPT-4 \cite{openai2023gpt4}, there are open-source LLMs available, such as LLaMA1/2 \cite{touvron2023llama, touvron2023llama2}. While these open-source models may not yet match the capabilities of their closed-source counterparts, they offer a significant advantage: the ability to run locally on a user's machine. This local execution ensures greater security for private data, making these models a more secure option for handling sensitive information.

The preceding overview delineates the particular NLP tasks essential for the automated resume screening framework. Additionally, it is highlighted that the tasks, as marked by the blue blocks in Figure \ref{1figure1}, are manageable through PLMs and LLMs. A succinct explanation of the fundamental principles of LMs is also provided. Subsequent paragraphs will offer a comprehensive exposition on the implementation of an automated resume screening system utilizing agents derived from LLMs.

\begin{figure}[!h]
\centering
\includegraphics[width=228 pt]{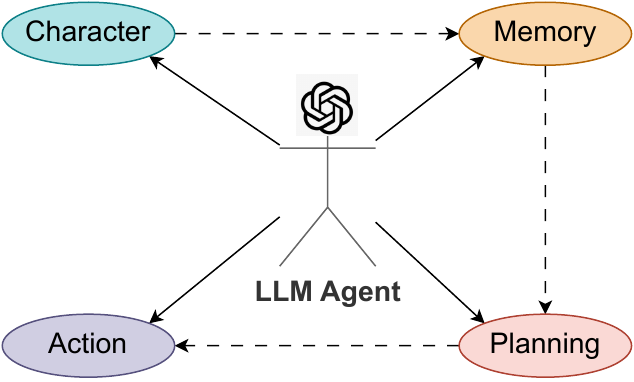}
\caption{\label{3figure3}The illustration depict LLM as the backbone of the agent system.}

\end{figure}

Figure \ref{3figure3} presents a schematic representation of a fundamental agent system. This diagram illustrates the segmentation of Language Model (LLM) agents into four core components: Character, Memory, Planning, and Action. Initially, the LLM agent is assigned a distinct character, essentially defining its role or function. For instance, in this study, the LLM agent is designated as an adept Human Resources (HR) professional. This role encapsulates the responsibilities and duties expected of the LLM agent. Subsequently, 'Memory' pertains to the requisite knowledge base necessary for the agent to execute its role effectively. In the context of an HR professional, this encompasses a comprehensive understanding of employee skill requirements, salary management, and relevant laws and regulations. This aspect is analogous to an LLM's capability to access and utilize its internal knowledge database. The next phase involves 'Planning,' where the LLM agent strategizes the execution of tasks. This process entails decomposing a complex task into smaller, manageable subtasks, thereby enhancing the efficiency in addressing intricate assignments. This stage is indicative of an LLM's reasoning and problem-solving abilities. Finally, the 'Action' component represents the implementation stage. In the context of an automated resume screening system, this would involve the LLM agent filtering and selecting resumes that align with specific job requirements. This final stage exemplifies the practical application of the LLM agent's planning and reasoning in a real-world scenario.

In this study, we integrate a LLM agent into the process of automated resume screening. We propose an innovative framework that leverages the LLM agent for automated extraction and analysis of resumes. This framework streamlines the entire process, from initial resume screening to the final selection of qualified candidates, significantly enhancing the efficiency of this task. For our analysis, we utilized a publicly available IT industry-specific resume 
dataset\footnote{\url{https://huggingface.co/datasets/ganchengguang/resume_seven_class}}, optimized for sentence classification. Through fine-tuning of the LLM, we achieved an F1 score of 87.73 in sentence classification. This improvement is particularly notable in the model's ability to identify and exclude personal information from resumes, thereby mitigating the risk of privacy breaches when employing models like GPT-3.5/4. Additionally, we developed an HR Agent, designed to both grade and summarize resumes. We created a specialized Grade \& Summarization Resume (GSR) dataset, derived from the initial dataset, using the GPT-4 model. This GSR dataset was instrumental in evaluating other LLMs. In these evaluations, the LLaMA2-13B model, once fine-tuned, achieved a ROUGE-1 score of 37.30 in summarization and a Grade accuracy of 81.35, significantly surpassing the baseline GPT-3.5-Turbo model. Finally, we deployed the HR Agent to select suitable candidates, further analyzing the decision-making outcomes.

In addition, we conducted experiments using GPT-4-Turbo and GPT-3.5-Turbo-16k to demonstrate that LLMs are capable of processing long-context resume information effectively. To further validate the effectiveness of our proposed LLM-based resume screening framework, we randomly selected 50 resumes for manual summarization and evaluation. The performance of the LLMs was benchmarked against this manually labeled dataset. Our analysis of the experiments and specific samples indicated that LLMs' evaluations and decisions closely resemble those of human reviewers. Additionally, to assess the framework's ability to meet complex recruitment requirements, we incorporated additional criteria beyond the basic requirements into the framework. The decision-making outcomes were then analyzed to determine the adaptability of the LLMs to these enhanced requirements.

Our comprehensive experiments and analysis demonstrate the LLM agent's robust capability in resume screening. As an HR agent, it effectively facilitates the candidate selection process.

\section{Related Work}


\subsection{Resume Information Extraction}

Resume screening is a classic application of information extraction, evolving from rule-based methods \cite{mooney1999relational} to the use of toolkits for automating these rules \cite{ciravegna2004learningpinocchio}. Over time, techniques such as Hidden Markov Models (HMM) and Support Vector Machines (SVM) developed into Cascaded Hybrid Models for segment classification in resumes \cite{yu2005resume}. The adoption of deep learning, utilizing Convolutional Neural Networks (CNNs) and Long Short-Term Memory networks (LSTMs), further enhanced extraction methods \cite{harsha2022automated, sinha2021resume, kinge2022resume, ali2022resume, bharadwaj2022resume, zu2019resume, barducci2022end}, with Conditional Random Fields (CRFs) improving LSTM models by refining sequence labeling \cite{ayishathahira2018combination}.

Recent advances incorporate pre-trained language models like BERT, integrated with LSTMs and CRFs, significantly enhancing contextual understanding for resume information extraction \cite{tallapragada2023improved}. This has been applied in developing algorithms for automating recruitment, with applications in ranking candidates for specific jobs \cite{10286578}.

Additionally, new tools such as PROSPECT have been developed to support resume screening by extracting and ranking candidate skills and experiences using CRFs \cite{singh2010prospect}. Another approach involves using NLP and similarity measures to improve the efficiency of job candidate selection through automated systems that match resumes with job descriptions \cite{daryani2020automated}.

\subsection{Large Language Model in Recruit Application}

After the advent of LLM, there were other jobs that used LLM in the recruitment process. The work \cite{du2024enhancing} introduces an LLM-based GANs Interactive Recommendation (LGIR) method that enhances job recommendation systems by using Generative Adversarial Networks to refine resume representations, improving the accuracy of job matching by overcoming issues of fabricated content and insufficient data. JobRecoGPT \cite{ghosh2023jobrecogpt} explores four job recommendation methods using LLMs to analyze unstructured job and candidate data, highlighting advantages, limitations, and efficiency in IT domain job matching.

\subsection{Decision Making with LLM Agent}

In addition, the LLM agent is employed in decision-making processes across various applications.
This paper \cite{huang2024far} evaluates the decision-making capabilities of LLMs in complex multi-agent environments using a novel framework. This paper \cite{ma2024towards} introduces a novel framework, Human-AI Deliberation, designed to enhance AI-assisted decision-making by fostering a deliberative dialogue between humans and AI. \cite{chen2023introspective} introduces "Introspective Tips," a novel approach for enhancing the decision-making capabilities of LLMs without the need for fine-tuning. \cite{wei2022chain} highlights that enhanced decision-making abilities can be achieved by incorporating a series of intermediate reasoning steps. \cite{yao2022react} presents ReAct, a novel method that integrates reasoning with action generation, enhancing the synergy between language comprehension and decision-making in interactive tasks.

\begin{figure*}[!t]
\centering
\includegraphics[width=456 pt]{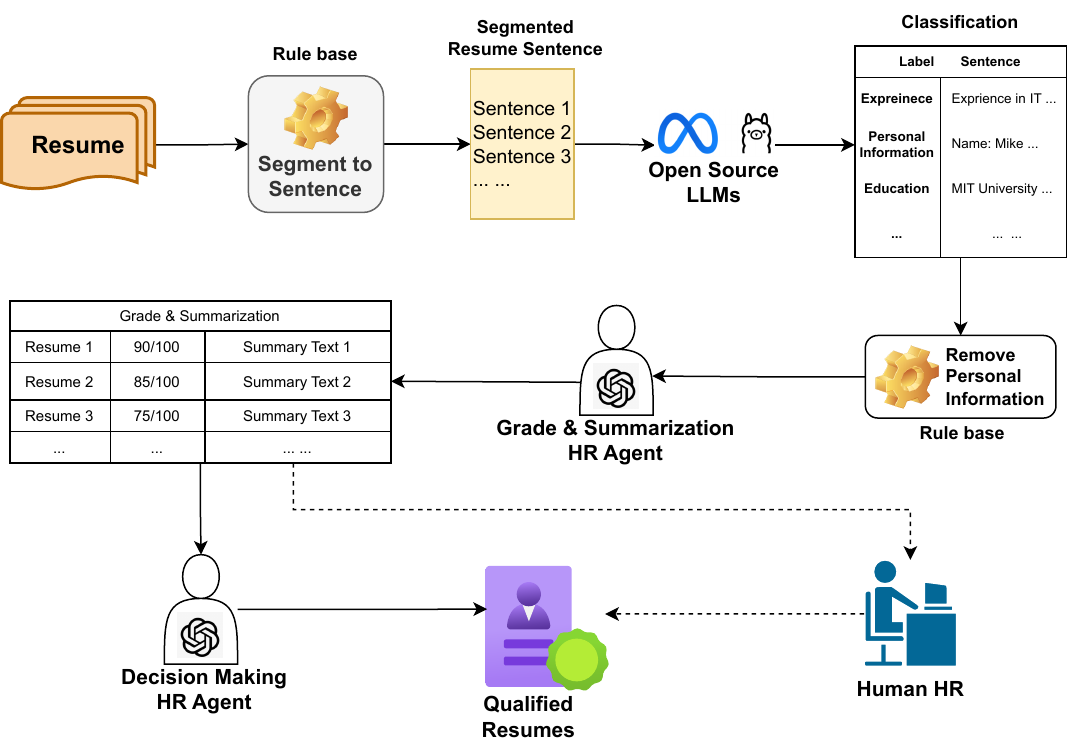}
\caption{\label{4figure4}The illustration depict the workflow of LLM agent base Automated Resume Screening Framework.}

\end{figure*}

\subsection{Compare LLM-based Resume Screening and Traditional Methods}

The application of LLMs to resume screening frameworks offers significant advantages over traditional methods. Firstly, unlike PLMs which are constrained to processing a maximum of 512 tokens, LLMs can manage considerably longer texts. This capability allows LLMs to effectively handle resumes of virtually any length, enhancing the comprehensiveness of the screening process. Secondly, LLMs possess a broader knowledge base, enabling their deployment across various industries for resume data processing without the need for specific fine-tuning. Furthermore, LLMs demonstrate enhanced performance compared to traditional PLMs, providing evaluations and judgments that are more aligned with human reasoning. This makes LLMs particularly valuable in contexts where nuanced understanding and decision-making are crucial.

\section{Resume Screening Framework Based on LLM Agents}

This section provides a comprehensive overview of the workflow within an novel automated resume screening framework that utilizes a LLM agent. It focuses on the application of the LLM agent in efficiently identifying and selecting qualified resumes from a substantial pool of candidates. To maintain clarity, this overview condenses some aspects, retaining only the essential steps. Detailed discussions of these steps are presented in the subsequent three subsections.

Figure \ref{4figure4} illustrates the architecture of our innovative automated resume screening system, which is underpinned by a LLM agent. The process begins with the transformation of a multitude of resumes, each in disparate formats like PDF, DOCX, and TXT, into a uniform JSON format. This is achieved through a rule-based algorithm designed to standardize the diverse formatting and file types into coherent, individual sentences. Such pre-processing is crucial for enabling consistent analysis in later stages. The next step involves segmenting these uniformly formatted resumes into distinct sentences, based on criteria like line breaks. This segmentation is vital for the effective functioning of the open-source LLM, which operates locally to classify each sentence. Critical to this process is the categorization of various sentence types, ranging from personal information, which is earmarked for removal to protect privacy, to other categories like work experience, education, and skills. This categorization is particularly significant because it allows for a tailored analysis based on the specific requirements of a job position. For instance, certain roles may prioritize a candidate's skills over their educational background. By extracting and focusing on the segments of a resume that detail relevant skills, the system can more effectively screen candidates for such positions. While our framework currently focuses primarily on the basic functionality of removing personal information, it lays the groundwork for more nuanced and customized resume screening processes in the future.

Upon removed personal information from resumes, the next step involves utilizing the GPT-3.5 model for grading and summarizing these documents. This task primarily falls under the purview of the HR agent. The grading system serves as a mechanism to rank candidates, streamlining the process of identifying top applicants. Summarization, on the other hand, is aimed at conserving time for the decision-making agent, who must evaluate these summaries. The brevity of summarized content not only expedites the process but also benefits human HR professionals by reducing the time required for initial resume screening. Once resumes are assigned grades and summaries, the decision regarding the candidates' progression can be made either by an HR agent or a human HR professional. Utilizing grades as a comprehensive metric allows for an efficient ranking of candidates. Depending on the specific requirements, a selection of the top 10 or 100 candidates can be made for the next stage of the screening process. This step, whether performed by an HR agent or a human, significantly reduces the time and effort involved in decision-making. The final stage involves choosing candidates for interviews or extending job offers directly, based on the refined pool of qualified resumes. This method optimizes the recruitment process, ensuring efficiency and effectiveness in candidate selection.

The preceding section outlined the comprehensive procedure for automated resume screening utilizing open source LLM and LLM agents. Subsequent subsections will elaborate on the implementation of the three pivotal steps: sentence classification, grade \& summarization, and decision-making.

\begin{figure}[!h]
\centering
\includegraphics[width=160 pt]{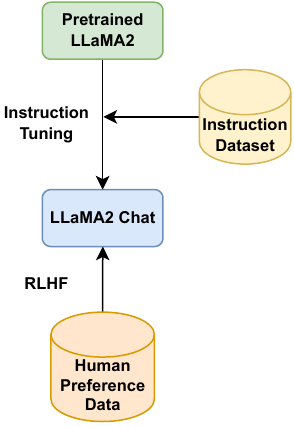}
\caption{\label{5figure5}The illustration depict the process of instruction tuning and RLHF for the LLaMA2 model.}

\end{figure}

\subsection{Sentence Classification}

In our methodology, the LLaMA2 model serves as the foundational base for sentence classification. We enhanced this base model through fine-tuning, specifically targeting the classification of resume sentences. Unlike previous Pretrained Language Models (PLMs), the LLaMA2 model does not straightforwardly accept a sentence as input and produce a corresponding predicted label. This limitation stems from the model's architecture, as depicted in Figure \ref{5figure5}. The LLaMA2-chat variant, developed from the original LLaMA2 model, undergoes a specialized instruction tuning process using an instruction dataset, followed by further refinement through Reinforcement Learning from Human Feedback (RLHF). This approach presents a challenge: simply inputting a sentence into the model does not guarantee the generation of the appropriate prediction label, a phenomenon also evidenced in our subsequent experimental results.

\begin{figure}[!t]
\centering
\includegraphics[width=200 pt]{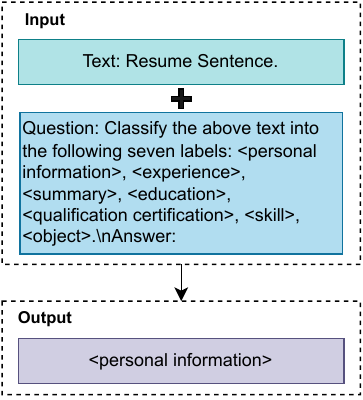}
\caption{\label{6figure6}The illustration depict the components of the converted resume sentence instruction dataset.}

\end{figure}

The underlying reason for this is the model's design to respond according to the instruction dataset's guidelines. To elaborate, the input not only contains the query sentence but also incorporates specific textual instructions guiding the model's response. As illustrated in Figure \ref{6figure6}, to address this, we append a question to the resume sentence requiring classification. This question instructs the model to categorize the preceding sentence into one of seven predefined labels. Alongside this, we introduce the "Answer:" prompt as part of the input text sequence. Consequently, we utilize the LLaMA2 model, fine-tuned with a specially curated resume sentence instruction dataset, for the effective classification of resume sentences. This fine-tuned LLaMA2 model demonstrates enhanced performance in the task at hand.

\subsection{Grade \& Summarization}

\begin{figure}[!h]
\centering
\includegraphics[width=220 pt]{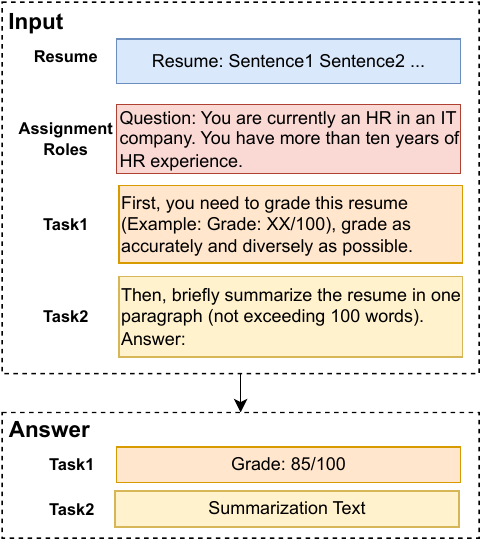}
\caption{\label{7figure7}The illustration depict assignment of roles and tasks to the LLM agent.}

\end{figure}

Upon extracting the resume text with personal details redacted, our objective is to assess and encapsulate each resume. This process involves a shared component: both evaluation and summarization require a comprehensive understanding of the resume's content. Consequently, we amalgamated these two processes into a singular question and answer task. Figure \ref{7figure7} illustrates this integration, where the red block denotes the assigned role to the LLM agent, exemplified as an HR professional in an IT firm with over a decade of HR experience. This role-play empowers the HR agent to conduct an analysis with the insight of a seasoned HR expert.

The initial task involves the HR agent appraising the resume, striving for precision and variety in assessment. For guidance, a scoring example (e.g., Grade: XX/100) is provided, deliberately without a predetermined score to avoid biasing the agent's evaluation. Following this, the agent is tasked with summarizing the resume in a concise paragraph, limited to 100 words. The culmination of this process is the agent presenting both the grade and a succinct summary of the resume.

\begin{figure}[!b]
\centering
\includegraphics[width=220 pt]{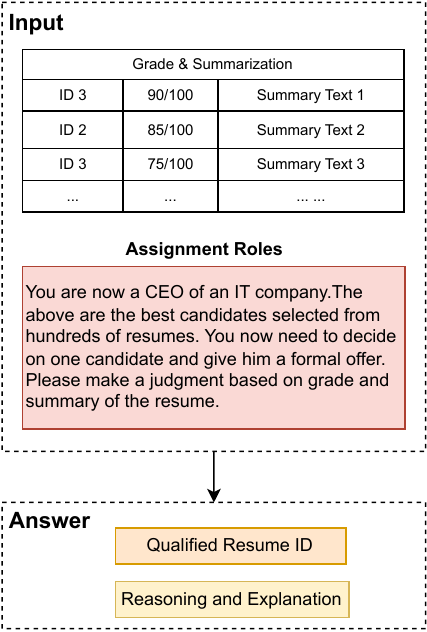}
\caption{\label{8figure8}The illustration depict the HR agent making a final Decision to select a qualified candidate.}

\end{figure}

\subsection{Decision Making}

The concluding phase of the resume screening system involves evaluating candidates based on their assigned grades and summaries. In this study, we have bifurcated this stage into two distinct processes: automatic and manual. This bifurcation allows for flexibility to cater to various requirements. Even when the ultimate selection is executed manually by human HR personnel, the highly-rated resumes can be efficiently sifted through utilizing grade rankings. Additionally, the provided summaries facilitate a rapid comprehension of the key elements in each resume by the HR staff, thereby significantly reducing the time required for resume screening.

On the other hand, the process of automated decision-making can be further pursued through the use of a LLM agent. As depicted in Figure \ref{8figure8}, each resume is initially provided with a formatted identifier, grade, and summary. This procedure simulates the selection of final candidates. Consequently, the role assignments in the red block are altered, transitioning from an experienced HR professional to a CEO. The task involves selecting one candidate out of ten, based on the provided grades and summaries. Following this, the agent will identify the chosen resume by its ID and articulate the rationale behind this particular selection.

Consequently, a multitude of resumes undergo a series of evaluative processes to identify the most suitable candidates. The automated resume screening framework employed in this process is versatile, allowing customization to meet various requirements and real-world scenarios. For instance, this research replicates the resume evaluation criteria of IT companies, which prioritize candidates' technical skills. Accordingly, the screening process emphasizes skill-related information in the resumes. This approach is adaptable to other sectors such as Marketing, Education, Finance, etc., by modifying the keywords and criteria. Furthermore, the system can be designed to mitigate educational bias by prioritizing skills and work experience, thus focusing on the candidates' competencies. Additionally, the framework's screening parameters are flexible; for example, it can be set to select the top 10\% of candidates based on specific criteria. In summary, this adaptability enhances the overall effectiveness and applicability of the screening framework.

\section{Experiment Setup}

In this section, we will introduce how to simulate a resume screening process to verify the effectiveness of the automated resume screening framework based on LLM agent.
This includes the preparation of the resume dataset and some settings for simulating the resume screening \ref{resumedataset}. The selection of LLM for the backbone of the LLM agent, and the parameter settings for model inference and fine-tuning \ref{preparellm}. And description of the evaluation method \ref{evaluationmetric}.

\subsection{Resume Dataset and Screening Simulation}\label{resumedataset}

In the initial phase of our study, we opted for a classification dataset comprising sentences from resumes \cite{gan2022construction}. This dataset encompasses seven categories: personal information, experience, summary, education, qualification certification, skill, and objectives. It includes a total of 1,000 resumes, amounting to 78,668 sentences, predominantly from the IT sector. Thus, the simulation of resume screening in this research is contextualized within an IT company recruitment framework. And we set that the person who is used to grade each resume is an experienced HR stuff. Then, we set that the top 10 resumes of grade go to the final round of decision making. Finally, the CEO is set to screen the resume grades and summaries of these 10 candidates in order to select a final qualified candidate.

Conversely, given the lack of grade and summarization annotations in the original resume dataset, the GPT-4 model, which currently exhibits superior performance, was employed for annotating these resumes. The annotations generated by GPT-4 served as a benchmark for evaluating the performance of other models, essentially treating GPT-4's output as a gold standard (100\% performance) against which to measure other LLMs. This approach facilitated the creation of a comprehensive dataset for simulating resume screening processes. Moreover, due to the token limit of 4096 in the LLaMA2 model, resumes exceeding this token count were excluded. Consequently, a refined dataset of 838 resumes remained, which was then utilized for the second phase of testing.


To enhance the validation of our proposed resume screening framework, we randomly selected 50 resumes, which were then summarized and evaluated manually. This process mirrored the previous method of labeling using GPT-4, where each resume was concisely summarized in approximately 100 words and assessed on a 100-point scale.

We enlisted three graduate students to annotate the resumes manually. Before beginning the annotation process, these evaluators received comprehensive training and were provided with several exemplars to standardize their markings. Specifically, the summaries required detailed inclusion of the candidate's work experience, years in the field, educational achievements (including undergraduate and graduate degrees), skills, experience at major companies, and any other notable experiences, while adhering strictly to 100 word limit.

During the grading phase, we establish specific criteria for evaluation. For instance, we consider skills that may not be directly relevant to the needs of an IT company, such as marketing management. Candidates with limited work experience typically receive grades between 50 and 65. Conversely, candidates who possess several years of IT experience along with undergraduate and graduate degrees in computer science are usually scored within the range of 80 to 95. Due to the inherent imprecision of the scoring process, we adopt a scoring interval of 5 points. Ultimately, the grades are averaged across three evaluators. We then review three different summaries of each resume and select the one that most accurately reflects the original document as the final labeled result.

\subsection{Prepare Backbone LLMs and Parameter Sets}\label{preparellm}

In the initial phase of the sentence classification task, the LLaMA2-7B model was chosen for fine-tuning. The dataset, comprising 78,668 sentences, was partitioned into training, validation, and testing sets in a 7:1.5:1.5 ratio. A random seed of 42 was set to ensure reproducibility. This configuration aligns with the experimental setup described in the original paper pertaining to the resume dataset, enabling direct comparisons with other PLMs. For the training process, each GPU was assigned a batch size of 32, and the model underwent training for 2 epochs using 32-bit floating-point precision.

In the subsequent phase, specifically the second stage of grading and summarization, we selected LLaMA2-7B/13/70B and GPT-3.5-turbo-0614 as the backbone LLMs for the HR agent. Initially, we employed a zero-shot methodology to grade and summarize 838 resumes using four different LLMs, aiming to assess and compare their efficacy. During this process, we meticulously configured the parameters for model generation. The maximum number of new tokens was set at 200. This parameter choice was informed by the requirement that each resume should be graded and summarized in over 100 words. Additionally, we incorporated the 'do sample' and 'early stopping' features to optimize the summarization process. Except for these specific adjustments, all other parameters were maintained at their default settings.

In additional, we involved enhancing LLaMA2-7B/13B's capabilities by fine-tuning it with a specialized dataset focused on resume grading and summarization. Initially, this dataset was partitioned into two distinct subsets: a training set with 500 resumes and a test set comprising 383 resumes. Subsequently, the model underwent a training process where each GPU was allocated a batch size of eight. This training was conducted over 2 epochs, utilizing BF16 precision to optimize performance and computational efficiency.

In conclusion, our experimental setup involved conducting the inference tests for LLaMA2-7B/13B using a dual RTX 3090 24G GPU configuration with float16 precision. In contrast, both the fine-tuning procedures for LLaMA2-7B/13B and the inference tests for LLaMA2-70B were executed on an RTX A800 80G * 8 GPU server.

\subsection{Evaluation}\label{evaluationmetric}

In the initial phase of resume sentence classification, we utilize the F1 score as the primary evaluation metric. This score comprehensively reflects the model's performance by harmonizing precision and recall into a balanced mean. This approach offers a more accurate representation of the model's effectiveness.

For the resume summarization segment, our evaluation employs two predominant metrics: ROUGE-1/2/L \cite{lin2004automatic} and BLEU. These metrics are extensively recognized in the automatic evaluation of summarization tasks. Although BLEU is traditionally associated with translation evaluations, its application in summarization tasks provides valuable insights. By incorporating BLEU, we aim to achieve a more holistic assessment of the summarization quality.

Regarding the evaluation of grade scores, our methodology focuses on accuracy. This is particularly crucial given the significant variance in grade distribution across different models. We adopt a tolerance range approach in calculating accuracy: a generated grade is deemed accurate if it falls within a margin of ±5 from the actual grade. The calculation adheres to the following principle: if the absolute difference between the predicted and the actual grade is 5 or less, the prediction is considered correct (recorded as 1, with 0 indicating an error). To derive the final grade accuracy, we divide the total count of correct predictions by the total number of actual grades (PG is denote Predict Grade, TG is denote True Grade).

\begin{equation*}
    \text{Accuracy} = \frac{\sum_{i=1}^{N} \mathbf{1}\left(\left|\text{PG}_i - \text{TG}_i\right| \leq 5\right)}{N}
\end{equation*}

\begin{table*}[!h]
\caption{Results of resume sentence classification dataset.}
\label{table7-1}
\hbox to\hsize{\hfil
\begin{tabular}{l|c}\hline
Model &F1 Score\\
\midrule
BERT Large & 86.67 \\
ALBERT Large & 86.40 \\
RoBERTa Large & 87.00 \\
T5 Large &	87.35 \\
LLaMA2-7B-chat &	78.16  \\
LLaMA2-7B-chat (Instruction Format) &	\textbf{87.73}  \\
\hline
\end{tabular}\hfil}
\end{table*}

\begin{table*}[!h]
\caption{Results of resume grade and summarization dataset (ROUGE-1/2/L).}
\label{table7-2}
\hbox to\hsize{\hfil
\begin{tabular}{l|ccc}\hline
Model &ROUGE-1 & ROUGE-2 & ROUGE-L\\
\midrule
LLaMA2-7B & 26.35 & 6.22 &  24.00 \\
LLaMA2-13B & 25.31 & 5.83 &  22.99  \\
LLaMA2-70B& 28.12 & 7.70 & 25.68   \\
GPT-3.5-Turbo &	\textbf{34.75} & \textbf{12.34} & \textbf{31.92}   \\

\hline
\end{tabular}\hfil}
\end{table*}

\begin{table}[!h]
\caption{Results of resume grade and summarization dataset (BLEU and Grade Accuracy).}
\label{table7-3}
\hbox to\hsize{\hfil
\begin{tabular}{l|cc}\hline
Model & BLEU & Grade Accuracy \\
\midrule
LLaMA2-7B & 2.66 & 47.49\\
LLaMA2-13B & 2.56 & \textbf{59.31}   \\
LLaMA2-70B& 3.73 & 23.27   \\
GPT-3.5-Turbo & \textbf{7.31} & 47.61   \\

\hline
\end{tabular}\hfil}
\end{table}

\section{Results}

In the results of sentence classification for resumes, we conducted comparative experiments on the performances of several large-scale models: BERT Large, ALBERT Large, RoBERTa Large, and T5 Large. The results, detailed in Table \ref{table7-1}, reveal a notable enhancement in the F1 score of the LLaMA2-7B-chat model, which reaches 87.73, attributed to the implementation of the instruction format for both input and output. Interestingly, a direct fine-tuning of the LLaMA2-7B-chat model, using the conventional approach of inputting sentences and outputting labels as done with previous PLMs, resulted in a significant drop in the F1 score to 78.16. This outcome undergrades the efficacy of the instruction format we proposed. Furthermore, it highlights a critical consideration for fine-tuning LLMs in sentence classification tasks: adhering to the instruction format used during the instruction learning phase is crucial for optimizing the models' sentence classification capabilities.

In the evaluation of the grading and summarization component of the automated resume screening framework, we conducted tests using three different model sizes of LLaMA2 and GPT-3.5-Turbo. The results, as presented in Table \ref{table7-2}, indicate that GPT-3.5-Turbo outperformed the others across all three ROUGE metrics: ROUGE-1 (34.75), ROUGE-2 (12.34), and ROUGE-L (31.92), significantly surpassing the LLaMA2-70B model. Furthermore, under the BLEU evaluation metric (Table \ref{table7-3}), GPT-3.5-Turbo achieved a score of 7.31, nearly tripling the performance of its counterparts. This suggests that, if not using the fine-tuning method (0-shot inference). Utilizing closed-source models like GPT-3.5-Turbo and GPT-4 as the backbone for HR agents is crucial for enhanced performance. Interestingly, in the aspect of grading accuracy, LLaMA2-13B outshined the other models with a score of 59.31, notably exceeding the LLaMA2-70B model by 23.27. This anomaly and its implications will be further analyzed and discussed in the following subsection.

\begin{table*}[!h]
\caption{Results of fine-tuned LLaMA2-7B/13B in resume grade and summarization dataset (ROUGE-1/2/L).}
\label{table7-4}
\hbox to\hsize{\hfil
\begin{tabular}{l|ccc}\hline
Model &ROUGE-1 & ROUGE-2 & ROUGE-L\\
\midrule
GPT-3.5-Turbo & 34.61 & 12.18 &  31.83 \\
LLaMA2-7B & 36.50 & 13.32 &  33.48 \\
LLaMA2-13B &	\textbf{37.30} & \textbf{13.90} & \textbf{33.93}   \\

\hline
\end{tabular}\hfil}
\end{table*}

\begin{table}[!h]
\caption{Results of fine-tuned LLaMA2-7B/13B in resume grade and summarization dataset (BLEU and Grade Accuracy).}
\label{table7-5}
\hbox to\hsize{\hfil
\begin{tabular}{l|cc}\hline
Model & BLEU & Grade Accuracy \\
\midrule
GPT-3.5-Turbo & 7.40 & 45.24\\
LLaMA2-7B & 8.45 &  76.19 \\
LLaMA2-13B& \textbf{8.62} & \textbf{81.35}   \\

\hline
\end{tabular}\hfil}
\end{table}

Finally, the LLaMA2-7B/13B model was subjected to fine-tuning, yielding notable improvements as documented in Table \ref{table7-4}. Specifically, the refined LLaMA2-13B model demonstrated remarkable grades of 37.30, 13.90, and 33.93 in ROUGE-1/2/L metrics, respectively. This performance notably surpassed that of the 0-shot GPT-3.5 Turbo model in the test set evaluations. Furthermore, Table \ref{table7-5} presents the enhancements in BLEU grades, where the LLaMA2-7B and LLaMA2-13B models recorded increments to 8.45 and 8.62, respectively. Correspondingly, there was a significant improvement in grade accuracy, reaching 76.19 and 81.35 for each model. These results clearly indicate that, with adequate resume datasets for fine-tuning, opting for open-source LLaMA2-7B/13B models as the foundation for HR agent systems is a more effective strategy.

\begin{figure*}[!t]
    \centering
    \subfloat[Grade Distribution of LLaMA2-7B]{%
        \includegraphics[width=0.31\textwidth]{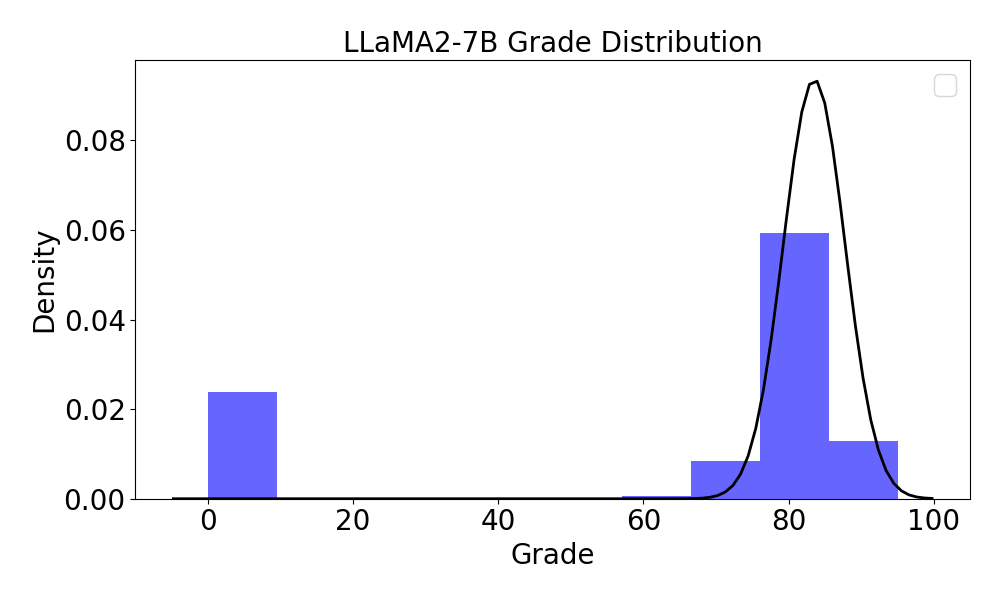}%
        \label{fig9:a}%
    }\hfill
    \subfloat[Grade Distribution of LLaMA2-13B]{%
        \includegraphics[width=0.31\textwidth]{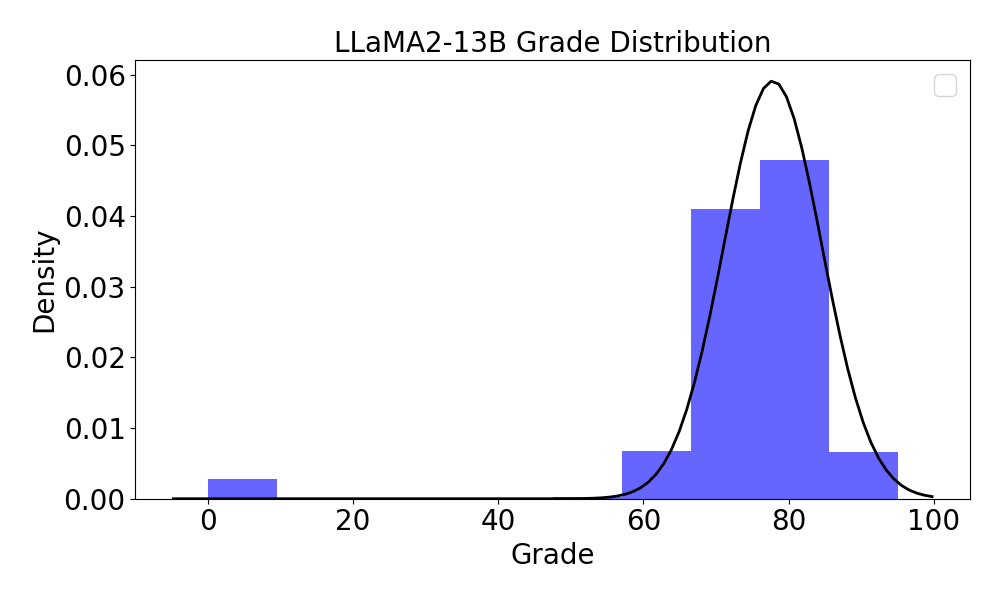}%
        \label{fig9:b}%
    }\hfill
    \subfloat[Grade Distribution of LLaMA2-70B]{%
        \includegraphics[width=0.31\textwidth]{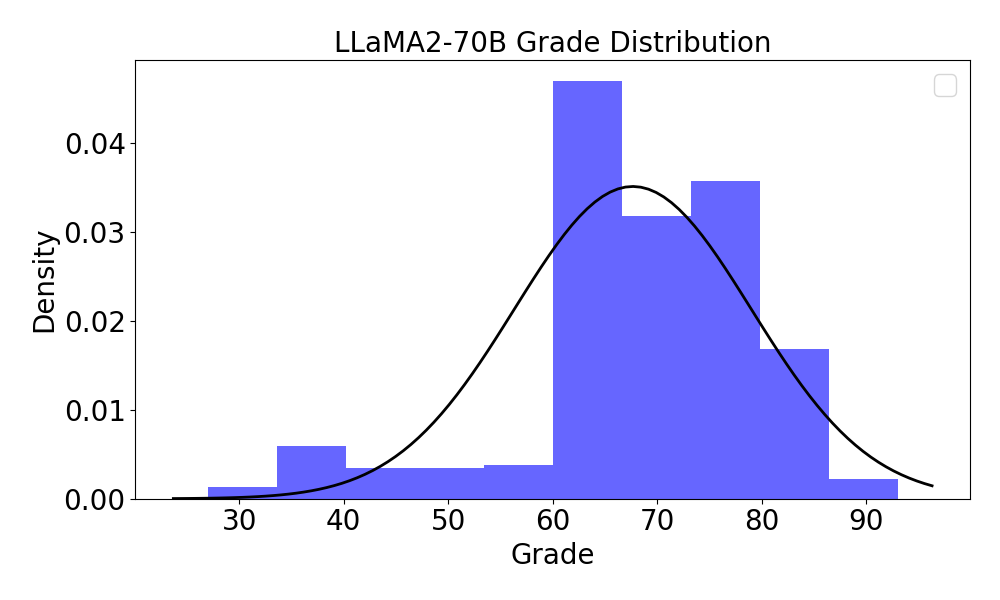}%
        \label{fig9:c}%
    }
    \caption{Compare the Grade Distribution of LLaMA2-7B/13B/70B models.
    }
    \label{fig9:above3}
\end{figure*}

\begin{figure*}[!t]
    \centering
    \subfloat[Grade Distribution of GPT-3.5-Turbo]{%
        \includegraphics[width=0.31\textwidth]{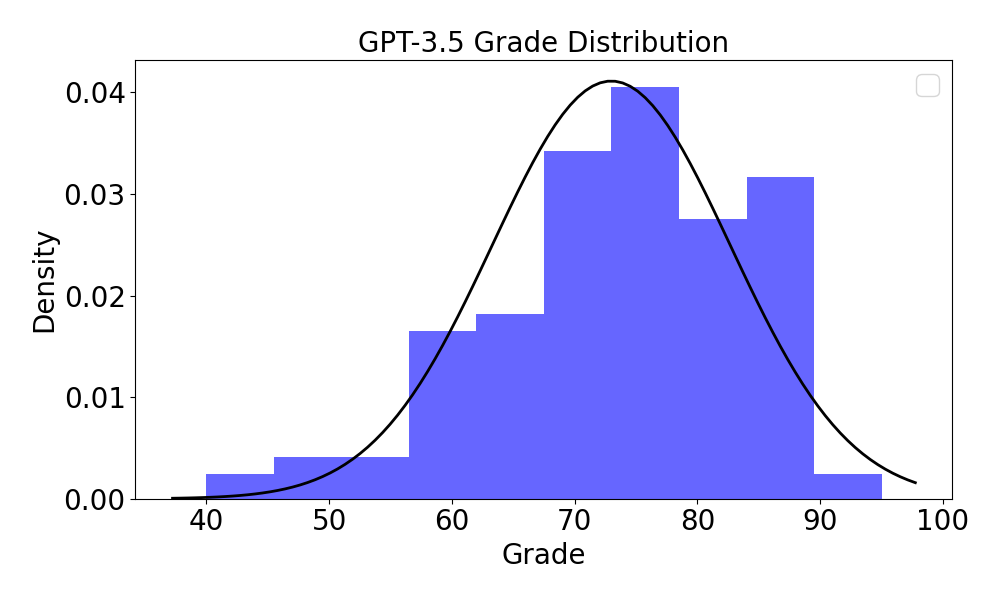}%
        \label{fig10:a}%
    }\hfill
    \subfloat[Grade Distribution of GPT-4]{%
        \includegraphics[width=0.31\textwidth]{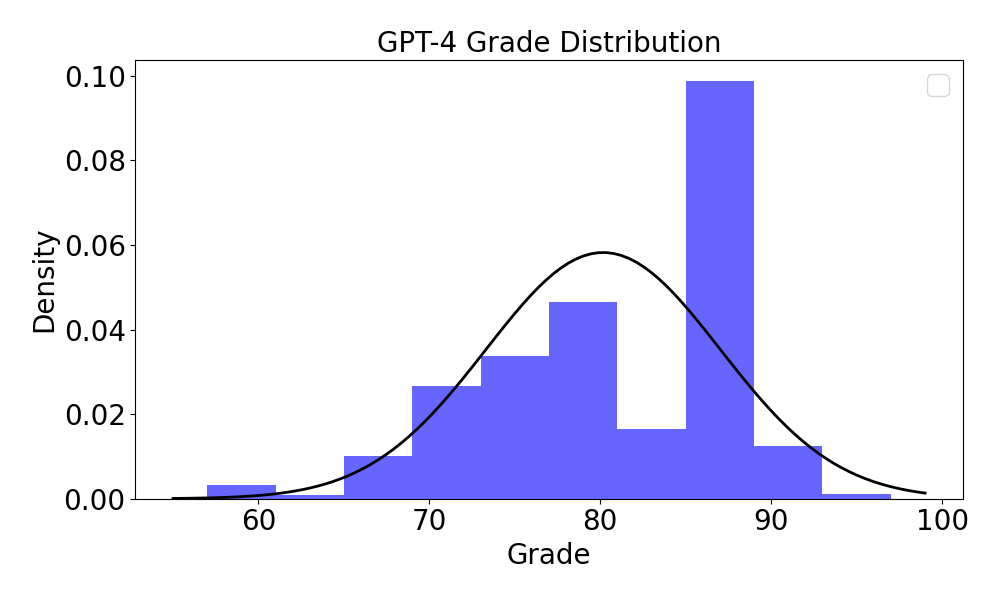}%
        \label{fig10:b}%
    }\hfill
    \subfloat[Comparison of 6 LLMs in grade]{%
        \includegraphics[width=0.31\textwidth]{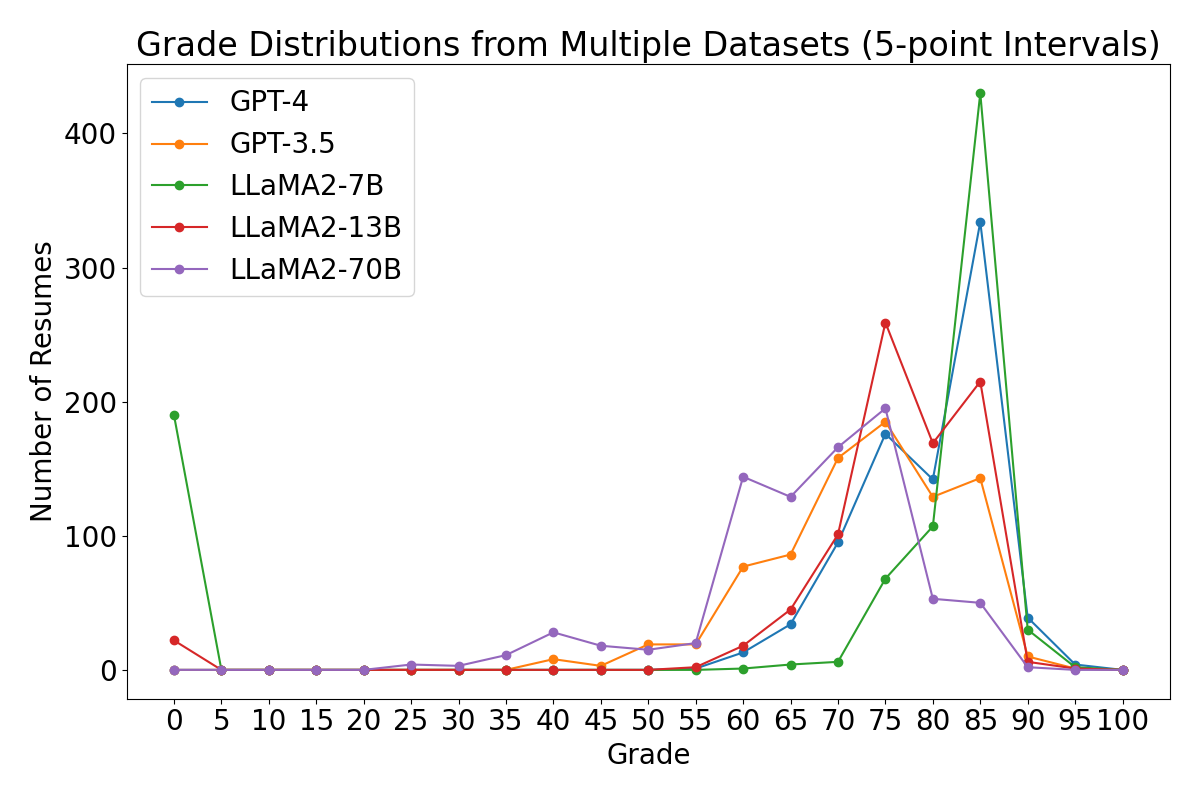}%
        \label{fig10:c}%
    }
    \caption{Compare the Grade Distribution of GPT-3.5-Turbo/4 models. And comparison of 6 LLMs in grade.
    }
    \label{fig10:under3}
\end{figure*}


\subsection{Normal Distribution of Grade}

Figure \ref{fig9:above3} \& \ref{fig10:under3} presents the normal distribution plots for the evaluations assigned by five different LLMs. Notably, the GPT-4 model generally aligns with the normal distribution across all grades, with a marked preference for assigning grades within the 85-90 range. This skew towards higher grades may stem from GPT-4's inclination to award more favorable ratings during fine-tuning processes, such as RLHF. Despite this, the impact on final resume screening remains minimal, as the system consistently prioritizes the top 10 resumes based on grades. While there may be some uncertainty regarding the extent to which these LLM-based HR agents accurately reflect the actual quality of each resume, the simulation experiment suggests that the grading patterns of all five LLMs largely adhere to a normal distribution. This indicates that the application of LLMs in resume evaluation is a successful experiment, with outcomes mirroring those expected in real-world scenarios.

\begin{table}[!h]
\caption{Number of grading errors (The grade is not a two-digit number) by different LLMs.}
\label{table7-7}
\hbox to\hsize{\hfil
\begin{tabular}{l|cc}\hline
Model & Total Number of Errors \\
\midrule
LLaMA2-7B & 190 \\
LLaMA2-13B & 22  \\
LLaMA2-70B& 8  \\
LLaMA2-7B FT&  1 \\
LLaMA2-13B FT&  0 \\

\hline
\end{tabular}\hfil}
\end{table}

\begin{figure*}[!t]
\centering
\includegraphics[width=438 pt]{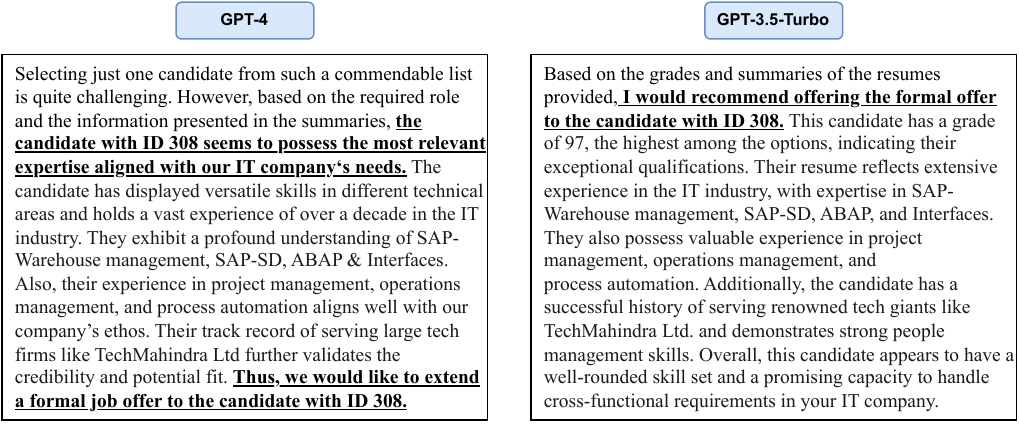}
\caption{The answer text of Decision Making with HR agents (GPT4 and GPT-3.5-Turbo Models).}
\label{figure6-17}
\end{figure*}

\begin{figure*}[!ht]
\centering
\includegraphics[width=348 pt]{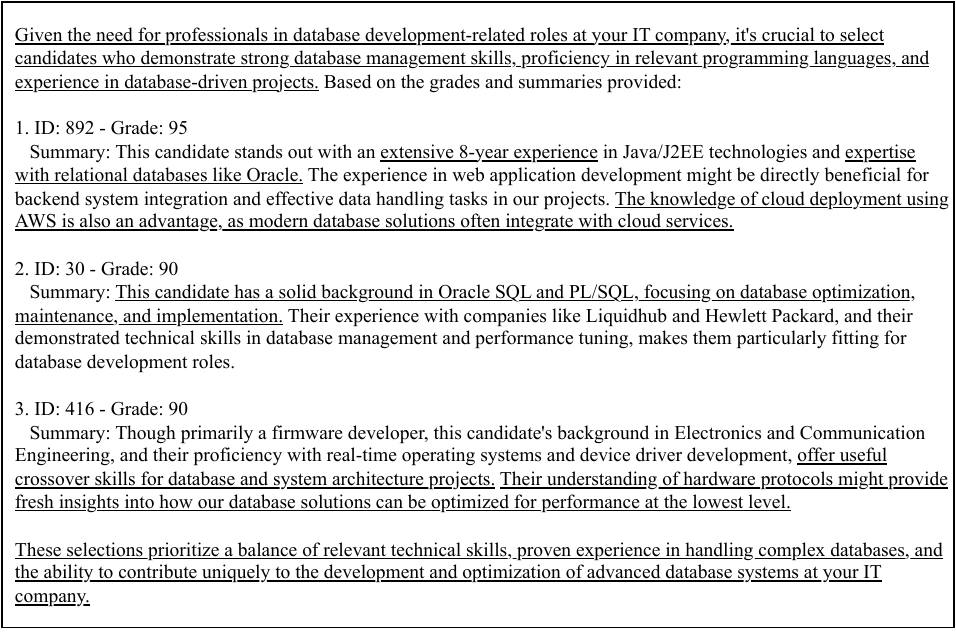}

\caption{The text of Decision Making with HR agents (GPT4-Turbo Models).}
\label{figure6-18}
\end{figure*}

The data presented in Figure \ref{fig9:above3} \& \ref{fig10:under3} and Table \ref{table7-7} reveals that the three LLaMA2 models exhibit instances of zero grading. This phenomenon occurs because these models assign grades that are not exclusively two-digit grades (such as 'A', 'B+++', etc.), leading to misclassification. Consequently, we have classified all such instances as zero grades. It is noteworthy that the incidence of grading errors in the LLaMA2 model is significantly reduced following fine-tuning. Additionally, the GPT-3.5-Turbo/4 model demonstrates an absence of grade errors, which can be attributed to the differences in the capabilities of various LLMs in terms of understanding and adherence to instructions.

\subsection{Analysis of Decision Making}

In our study, we utilized the GPT-3.5-Turbo and GPT-4 models as autonomous HR agents to evaluate the top 10 resumes based on their grades. The rationale behind their decisions is detailed. As illustrated in Figure \ref{figure6-17}, both models consistently identified resume ID 308 as the top candidate. The justification for this selection was not only the high grade of resume ID 308 but also its alignment with the specific needs of an IT company, including relevant work experience and managerial skills. This analysis demonstrates a remarkable congruence with the cognitive processes and judgment criteria typically employed by human HR professionals in decision-making. Furthermore, these findings underscore the potential of integrating LLM based HR agents into future automated resume screening systems.


To further investigate the decision-making capabilities of the HR agent, particularly in handling complex recruitment requirements, we refined the criteria within this stage and conducted an additional experiment. This experiment utilized a dataset of 50 manually annotated resumes, summarized and graded for relevance. We configured the hiring criteria to target three individuals with expertise in database development. This requirement was incorporated into the input prompt template as follows: "You are now recruiting three individuals for database development roles in your company."

As depicted in Figure \ref{figure6-18}, the HR agent successfully identified three candidates, providing detailed justifications for each selection. Notably, all candidates demonstrated relevant database development skills and substantial professional experience. The reasoning for their selection was well-articulated and convincing. Subsequent manual review of the candidates' resumes confirmed that these individuals were indeed the most suitable for the positions.

This experiment underscores the adaptability of the LLM-based resume screening framework, highlighting its ability to accommodate a diverse array of job specifications. It demonstrates that the model can be effectively tailored to meet varying recruitment needs of different companies for various positions, thus proving its generalizability and utility in complex HR scenarios.


\begin{table*}[!h]
\caption{Experimental results of LLMs evaluated based on manually annotated 50 sample datasets (ROUGE-1/2/L).}
\label{table7-8}
\hbox to\hsize{\hfil
\begin{tabular}{l|ccc}\hline
Model &ROUGE-1 & ROUGE-2 & ROUGE-L\\
\midrule
LLaMA2-7B & 27.03 & 7.11 & 24.28 \\
LLaMA2-13B & 24.96 & 5.96 & 22.62   \\
LLaMA2-70B & 27.27 & 7.69 & 25.00   \\
GPT-3.5-Turbo & 34.55 & 12.37 & 31.94  \\
GPT-4 & \textbf{39.87} & \textbf{16.44} & \textbf{35.89}   \\

\hline
\end{tabular}\hfil}
\end{table*}

\begin{table}[!h]
\caption{Experimental results of LLMs evaluated based on manually annotated 50 sample datasets (BLEU and Grade Accuracy).}
\label{table7-9}
\hbox to\hsize{\hfil
\begin{tabular}{l|cc}\hline
Model & BLEU & Grade Accuracy \\
\midrule
LLaMA2-7B & 3.28 & 22.00\\
LLaMA2-13B & 2.71 &  40.00  \\
LLaMA2-70B & 3.73 & 38.00   \\
GPT-3.5-Turbo & 7.16 &  \textbf{58.00}  \\
GPT-4 & \textbf{11.06} & 50.00   \\
\hline
\end{tabular}\hfil}
\end{table}


\begin{figure}[!ht]
\centering
\includegraphics[width=180 pt]{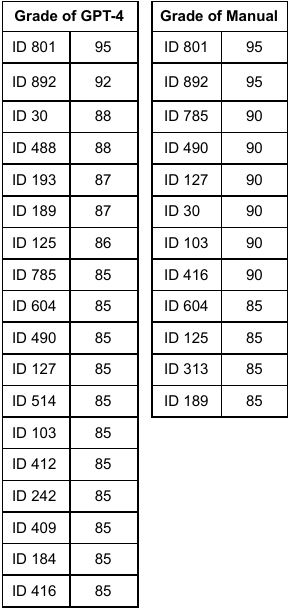}

\caption{Ranking comparison of top 10 GPT-4 rated resumes and top 10 manually graded resumes. Underlining represents the portion where the two overlap (i.e., the grades are equivalent).}
\label{table7-99}
\end{figure}

\subsection{Comparison with Manual Resume Screening}




We conducted a thorough evaluation of various LLMs by manually annotated 50 resumes to serve as a benchmark. The results of these tests are detailed in Tables \ref{table7-8} and \ref{table7-9}, with the GPT-3.5-Turbo and GPT-4 models demonstrating superior performance. Notably, while the accuracy of the grade assignments was not perfect, a subsequent analysis of the top ten resumes ranked by grades revealed significant insights. The resumes that received the highest grades from GPT-4 exhibited a striking resemblance to those scored manually, underscoring the effectiveness of the model in mimicking human evaluative patterns.

As depicted in Figure \ref{table7-99}, we compiled the IDs and grades of the top ten resumes according to the final grades from GPT-4 and manual scoring. In this figure, underlined text indicates where the two sets of rankings overlap, highlighting a strong correlation in the evaluation outcomes. Remarkably, both manually and by GPT-4, resumes ID 801 and ID 892 received the highest grades. Furthermore, 11 out of the 12 resumes that ranked highly in the manual assessment also featured prominently in the GPT-4 rankings, further validating the model's evaluative consistency.

Finally, we selected a final qualified resume using both manual and GPT-4 methods. Both selected resume ID 801 as the hiring candidate. Detailed analysis of this candidate's credentials revealed not only a robust six years of professional experience but also a comprehensive repertoire of IT-related skills. The individual is a versatile full-stack Java developer, proficient in a range of technologies spanning from front-end to back-end development, including networking. This skill set renders the candidate highly suitable for a developer role within an IT organization.

In conclusion, our findings affirm the efficacy of the proposed resume screening framework that leverages LLMs. This comparison with traditional manual methods substantiates the potential for LLMs to effectively replace manual resume screening processes in the future.



\begin{figure}[!h]
\centering
\includegraphics[width=218 pt]{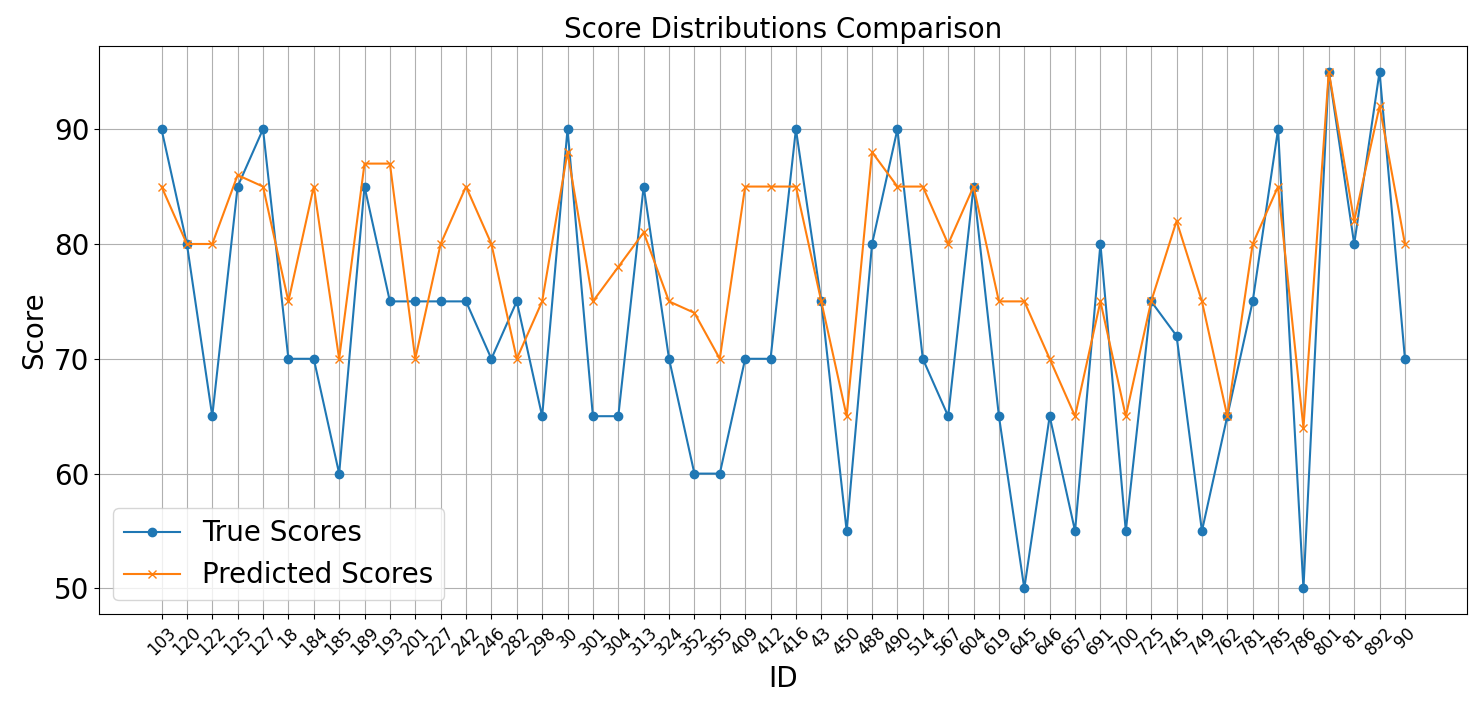}
\caption{The comparison of manual and GPT-4 in grades distributions (Base 50 samples dataset).}
\label{fig:gradeklcos}
\end{figure}

\begin{table*}[!h]
\caption{GPT-3.5-Turbo-16k Model experiment results. Evaluated based on GPT-4-Turbo (Max input length 128K) annotated 162 over length resume datasets .}
\label{table7-11}
\hbox to\hsize{\hfil
\begin{tabular}{l|ccc}\hline
Model &ROUGE-1 & ROUGE-2 & ROUGE-L\\
\midrule
GPT-3.5-Turbo-16k & 36.05 & 12.62 & 32.61 \\
\midrule
Model & BLEU & Grade Accuracy &   \\
\midrule
GPT-3.5-Turbo-16k & 6.78 & 72.22 &   \\

\hline
\end{tabular}\hfil}
\end{table*}

Our analysis also included a comparison between the score distributions of the most advanced GPT-4 model and manual grading. Figure \ref{fig:gradeklcos} illustrates this comparison, revealing a high degree of similarity between the two distributions. We quantified this similarity by calculating the cosine similarity, which yielded a value of \underline{\textbf{0.9944}}, approaching 1. This high similarity score further supports the consistency between GPT-4-generated grades and manual grades. This consistency is likely attributable to the model's use of instruction tuning and reinforcement learning with human feedback (RLHF). We also computed the correlation between the two rankings using Spearman's rho ($\rho$) and Kendall's tau ($\tau$). The values obtained were 0.7574 for Spearman's $\rho$ and 0.6252 for Kendall's $\tau$, indicating a strong positive correlation between the manual rankings and the predicted rankings produced by the LLM.

\subsection{Analysis of Long Length Resume Screening}

In addition, for resumes that exceed the LLaMA2 model's processing limit of 4,096 tokens, we conducted further experiments using more advanced models from the GPT family. Specifically, we utilized the GPT-4-Turbo and GPT-3.5-Turbo-16k models, which are capable of processing up to 128,000 and 16,000 tokens, respectively. These models are well-suited to handle the length of most resumes. Due to resource limitations, our experiments were confined to 162 resumes that exceeded 4,000 tokens in length.

We used the results from the GPT-4-Turbo model as a benchmark for evaluating the performance of the GPT-3.5-Turbo-16k model. As indicated in Table \ref{table7-11}, the GPT-3.5-Turbo-16k model demonstrated promising results, with a notable grade accuracy of 72.22\%. This high level of accuracy can be attributed to the model's ability to effectively analyze content-rich resumes, which typically contain extensive text detailing numerous skills and work experiences. Common sense suggests that resumes with more detailed information about a candidate's skills and experiences are likely to score higher, indicating a potentially stronger candidate. This principle was affirmed by our findings, which showed a direct correlation between the depth of resume content and the accuracy of the model's grading.


\subsection{Time comparison between automated and human resume screening}

Our study entailed a meticulous time comparison of three distinct resume screening methods: Automated, Semi-Automated, and Manual. To this end, we deconstructed the automated screening process into three discrete stages: Classification, Grading \& Summarization, and Decision Making. We measured the time expenditure for each phase, culminating in an aggregate duration assessment. Notably, in the Classification stage, we accounted for the time span from initiation to conclusion of the inference process, excluding the fine-tuning duration. This approach mirrors the actual operational timeline of the automated screening framework. In the Decision Making stage, our focus was on the time required to evaluate the top ten resumes.

Additionally, we assessed the time investment for the semi-automated method, wherein human HR personnel undertake the final decision-making step, while preceding stages are managed by LLMs. For the manual screening conducted by Human HR, we based our calculations on the average adult reading speed of 238 words per minute, as indicated by survey literature \cite{brysbaert2019many}. Consequently, we deduced that reviewing all 838 resumes, encompassing a total of 442,047 words, would approximately take 31 hours (Please note that this is an estimated time, calculated based on the average human reading speed.).

\begin{table*}[!t]
\caption{Follow each step to compare the time consumed by automated and manual resume screening.}
\label{table7-12}
\hbox to\hsize{\hfil
\begin{tabular}{l|cccc}\toprule
Model & Classification & Grade \& Summary & Decision Making  \\
\midrule
GPT-4 API & 25 min (FT LLaMA2-7B) & 2 h 30 min & 0.4 min  \\
\midrule
LLM with &  &  &  \\
Estimated Human    &25 min (FT LLaMA2-7B)  & 2 h 30 min (GPT-4) & 22 min (Manual)   \\
Screening Time   &  &  &    \\
\midrule
Estimated    &  &  &  \\
Human Screening  & --- & --- & --- \\
 Time  &  &  &    \\
\bottomrule
\end{tabular}}\hfil
\end{table*}

\begin{table*}[!t]
\hbox to\hsize{\hfil
\begin{tabular}{l|ccc}\toprule
Model & Total Time & Multiple & Automatic or Manual  \\
\midrule
GPT-4 API & 2 h 55.4 min & x 11 & Automatic \\
\midrule
LLM with Estimated & 3 h 17 min & x 9 & Semi-automatic \\
  Human Screening Time     &  &  &    \\
\midrule
Estimated Human &  & x 1 & Manual \\
Screening Time   & 31 h   &  & \\
\bottomrule
\end{tabular}}\hfil
\end{table*}

Table \ref{table7-12} illustrates that the fully automated resume screening framework, utilizing an LLM agent, completes the entire process set in approximately 2 hours and 55 minutes. This efficiency represents a speed 11 times faster than manual resume screening. Additionally, the semi-automatic approach is 9 times quicker than the manual method. While this comparison may lack rigorous precision, as it does not account for the possibility that human HR personnel might not read every word in a resume to reach a decision, the significant time reduction observed with the automated framework underscores its high efficiency.

\section{Conclusion}

In this study, we explore the feasibility of using an LLM agent for automated resume screening. We propose an innovative framework for this purpose and validate it using a real-world resume dataset, as well as through simulation of the resume screening process. Our results, derived from a series of comparative tests and analyses, demonstrate that the LLM agent can effectively perform the role of a human HR professional in resume screening. Notably, in terms of time efficiency, the LLM agent significantly surpasses traditional manual screening methods.

This work is subject to certain limitations. Primarily, it employs a controlled experimental design to maximize result accuracy, which restricts the scope of application to basic requirements of LLMs agent within IT companies. Consequently, this approach does not account for the varied requirements of other industries. Additionally, the collection of resume data is challenging due to privacy concerns. In future work, we aim to gather a broader array of resumes from diverse industries to enhance the representativeness of our study and further refine the LLM resume screening framework.

\bibliography{custom}

\end{document}